\def\year{2020}\relax
\documentclass[letterpaper]{article} 
\usepackage{aaai20}  
\usepackage{times}  
\usepackage{helvet} 
\usepackage{courier}  
\usepackage[hyphens]{url}  
\usepackage{graphicx} 
\usepackage{graphicx}  
\usepackage{microtype}
\usepackage{graphicx}
\usepackage{amsmath}
\usepackage{pgfplots}
\usepackage{tabularx}
\usepackage{booktabs}
\usepackage[symbol]{footmisc}
\usepackage{amssymb}
\usepackage{pifont}

\usepackage[toc,page]{appendix}

\usepackage[skins,breakable]{tcolorbox}
\pgfplotsset{compat=1.14}
\newtcolorbox{promptbox}[1][]{enhanced jigsaw,breakable,pad at break=1mm,
  oversize,left=8mm,interior hidden,colframe=black,nobeforeafter=,#1}
  
\title{Induced Natural Language Rationales and Interleaved Markup Tokens Enable Extrapolation in Large Language Models}

\begin{document}

\author{Mirelle Bueno\\
University of Campinas\\\And
Carlos Gemmell\\
University of Glasgow\\\And
Jeffrey Dalton\\
University of Glasgow\\\AND
Roberto Lotufo\\
University of Campinas\\
NeuralMind\\\And
Rodrigo Nogueira\\
University of Campinas\\
NeuralMind\\}

%
%

\maketitle              
\begin{abstract}

The ability to extrapolate, i.e., to make predictions on sequences that are longer than those presented as training examples, is a challenging problem for current deep learning models. 
Recent work shows that this limitation persists in state-of-the-art Transformer-based models.
Most solutions to this problem use specific architectures or training methods that do not generalize to other tasks.
We demonstrate that large language models can succeed in extrapolation without modifying their architecture or training procedure.
Our experimental results show that generating step-by-step rationales and introducing marker tokens are both required for effective extrapolation. First, we induce a language model to produce step-by-step rationales before outputting the answer to effectively communicate the task to the model. However, as sequences become longer, we find that current models struggle to keep track of token positions. To address this issue, we interleave output tokens with markup tokens that act as explicit positional and counting symbols.
Our findings show how these two complementary approaches enable remarkable sequence extrapolation and highlight a limitation of current architectures to effectively generalize without explicit surface form guidance. Code available at \url{https://github.com/MirelleB/induced-rationales-markup-tokens}

\end{abstract}

%
%
%

\section{Introduction}
 
\begin{figure}
  \includegraphics[width=0.49\textwidth]{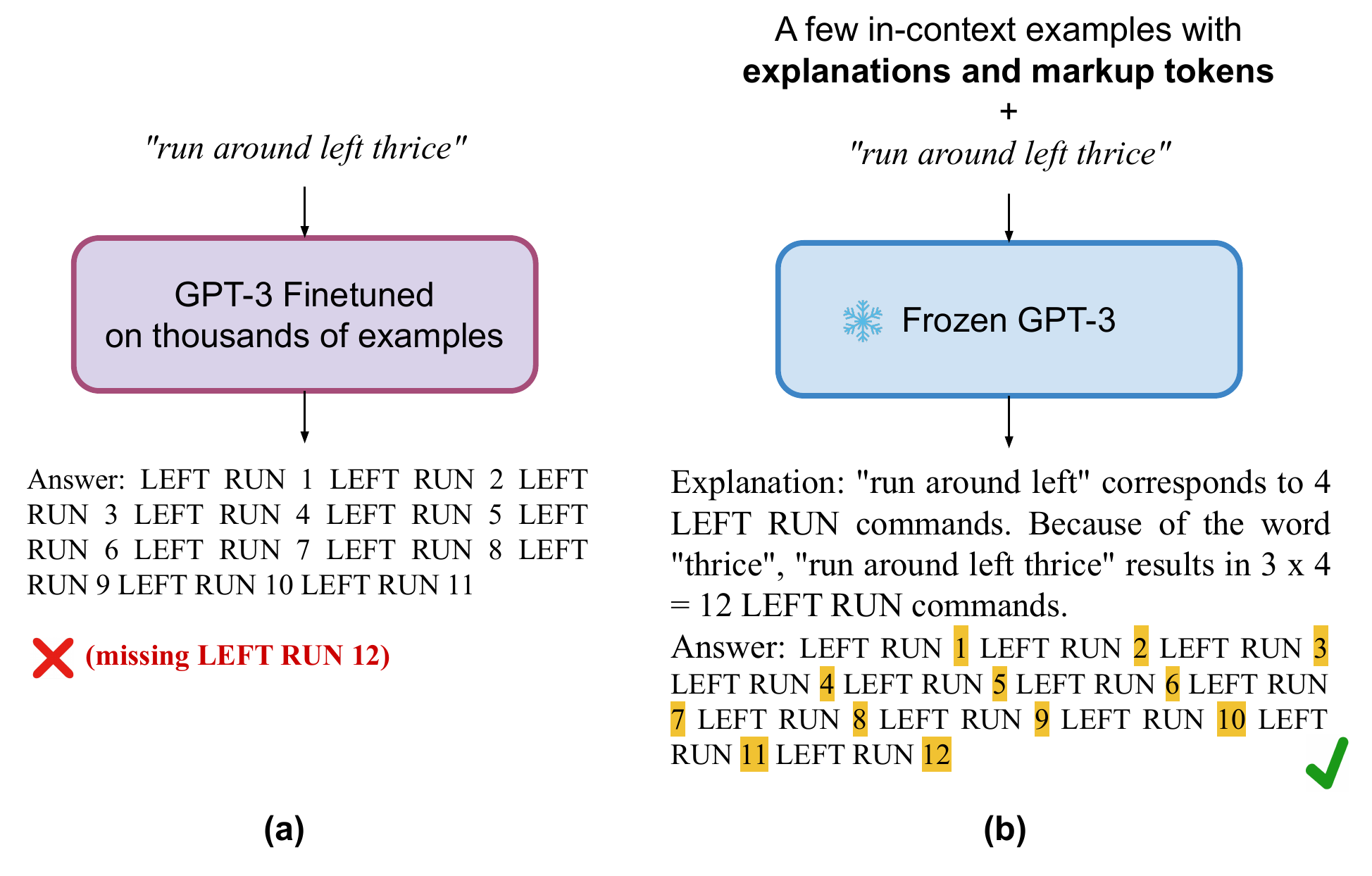}
  \caption{Answers produced by a GPT-3 model on the ``length'' split of the SCAN dataset when (a) fine-tuned on thousands of examples vs (b) induced via a few in-context examples to generate explanations and markup tokens (in yellow).}
  \label{fig:overview}
\end{figure}

The lack of compositional generalization of neural networks has been a long-standing limitation known for decades~\cite{fodor1988connectionism,Schmidhuber90towardscompositional,marcus1998rethinking,marcus2018deep,lake2018generalization,livska2018memorize,keysers2019measuring}. This is often associated with their failure to extrapolate, i.e., the ability to work on sequences that are longer than those presented as training examples.
Modern architectures such as the Transformer~\cite{vaswani2017attention}, which is the core component of state-of-the-art NLP models, perform poorly on this class of problems~\cite{bhattamishra2020ability,nogueira2021investigating,wang2021exploring,pal2021investigating,welleck2021symbolic,bogin2022unobserved,finlayson2022makes,DBLP:journals/corr/abs-2110-09419}.
In Figure~\ref{fig:overview}-(a), we illustrate how recent large language models such as GPT-3 fail at this task, even when fine-tuned on thousands of examples.

Architectures and training methods that target this specific problem are often developed based on synthetic tasks whose creation rules are known~\cite{das1992learning,li-etal-2019-compositional,russin2019compositional,andreas-2020-good,liu2020compositional,chen2020compositional,herzig-berant-2021-span,shaw2021compositional,zhu2021learning}. Thus, they resort to techniques such as augmenting the training data or biasing the model's architecture to internally represent these rules. However, improvements obtained on one compositional generalization benchmark do not transfer to others~\cite{furrer2020compositional}, i.e., they lose their ability to be used as competitive general-purpose models in real tasks, as these can seldom be solved with a small set of rules.

We study the behavior of Transformer models and demonstrate that this problem is not due to an intrinsic limitation of their training algorithm. We show that inducing autoregressive models to rationalize before making a prediction~\cite{wang2022self,zelikman2022star} is not enough to extrapolate on long sequences: to solve it, we introduce markup tokens~\cite{nogueira2021investigating,kim2021have}. The two general approaches together allow the models to achieve remarkable extrapolation generalization without requiring changes to the model or architecture.
These findings provide evidence that general-purpose models have the ability to both improve their effectiveness and interpretability \textit{at the same time}. The need to markup tokens also suggests there are fundamental issues that need to be addressed in the Transformer architecture, particularly the need for better positional representations. Thus, our study confirms and supports recent results from previous work that positional embeddings used in current state-of-the-art Transformer models cannot precisely track of token positions or perform precise counting~\cite{liu2020learning,thawani-etal-2021-representing,press2022train}.

\section{Related Work}

A long list of architectures and training methods attempt to improve the extrapolation capabilities of deep learning models. For instance, some are specifically designed to solve only a handful of tasks~\cite{singh1992transfer,kaiser2015neural,kalchbrenner2015grid,price2016extensions,andreas2016learning,andreas2017modular,trask2018neural}.
Pre-trained word embeddings find it difficult to extrapolate to unseen numbers in training \cite{wallace2019nlp}. Alternatives to improving the extrapolation ability of neural models include building neural models with a pre-training corpus of numerical text \cite{geva-etal-2020-injecting} or using scientific notation to represent numbers~\cite{zhang2020language}. Likewise, better numerical and compositional skills can be achieved by supplementing input texts with pre-computed numerical calculations \cite{andor2019giving} or explicitly assuming rules or mathematical equations from natural language texts ~\cite{liu2019tree,li2019modeling,zou2019quantity,zou2019text2math,shi2020sequence,qiu2021improving}.
Many of these models are capable of adding numbers larger than those seen during training. In contrast, more general-purpose architectures fail to extrapolate on numerical tasks \cite{joulin2015inferring,dehghani2018universal,schlag2019enhancing}.


Our work derives from recent findings that show that inducing the model to generate explanations in natural language leads to better performance in a wide variety of tasks~\cite{recchia2021teaching,fernandes2022learning,wang2022self,zelikman2022star,nye2022show,katz2022inferring,zhou2022least,khot2022decomposed}. In particular, the work proposed by ~\cite{zhou2022least} achieves state-of-the-art results in the extrapolation of tasks involving symbolic manipulation, compositional generalization and numerical reasoning.  Tasks are solved via few-shot learning applied to a large language model (e.g. text-davinci-002) in two main steps. The first step consists of reducing the question into sub-questions, then, in the second phase, a new interaction is made with the model, now solving sequentially the sub-questions generated in the previous step.

The results shown in Zhou et al.~\cite{zhou2022least} corroborate our intuition that explanations alone are not enough to achieve extrapolations. By inducing the model to generate explanations \textit{and} markup tokens, \textit{we provide evidence that compositional generalization can be achieved without sacrificing the general applicability on other tasks}, which is often a feature that is lost with architectural modifications.

However, a limitation of Zhou et al.'s and our method is that both require a programmatic post-processing step: Zhou et al. use a python script to convert the model output (e.g., \texttt{3*["LEFT"]}), which is in python notation, into the expected format of the final answer (e.g., \texttt{LEFT LEFT LEFT}); in our method, we programmatically remove the markup tokens from the final answer. We argue that the need to call an external script exposes a limitation in the current Transformer architecture, namely, that it cannot handle long sequences of repeated tokens.

\section{Methodology} 
\begin{figure}
  \includegraphics[width=0.49\textwidth]{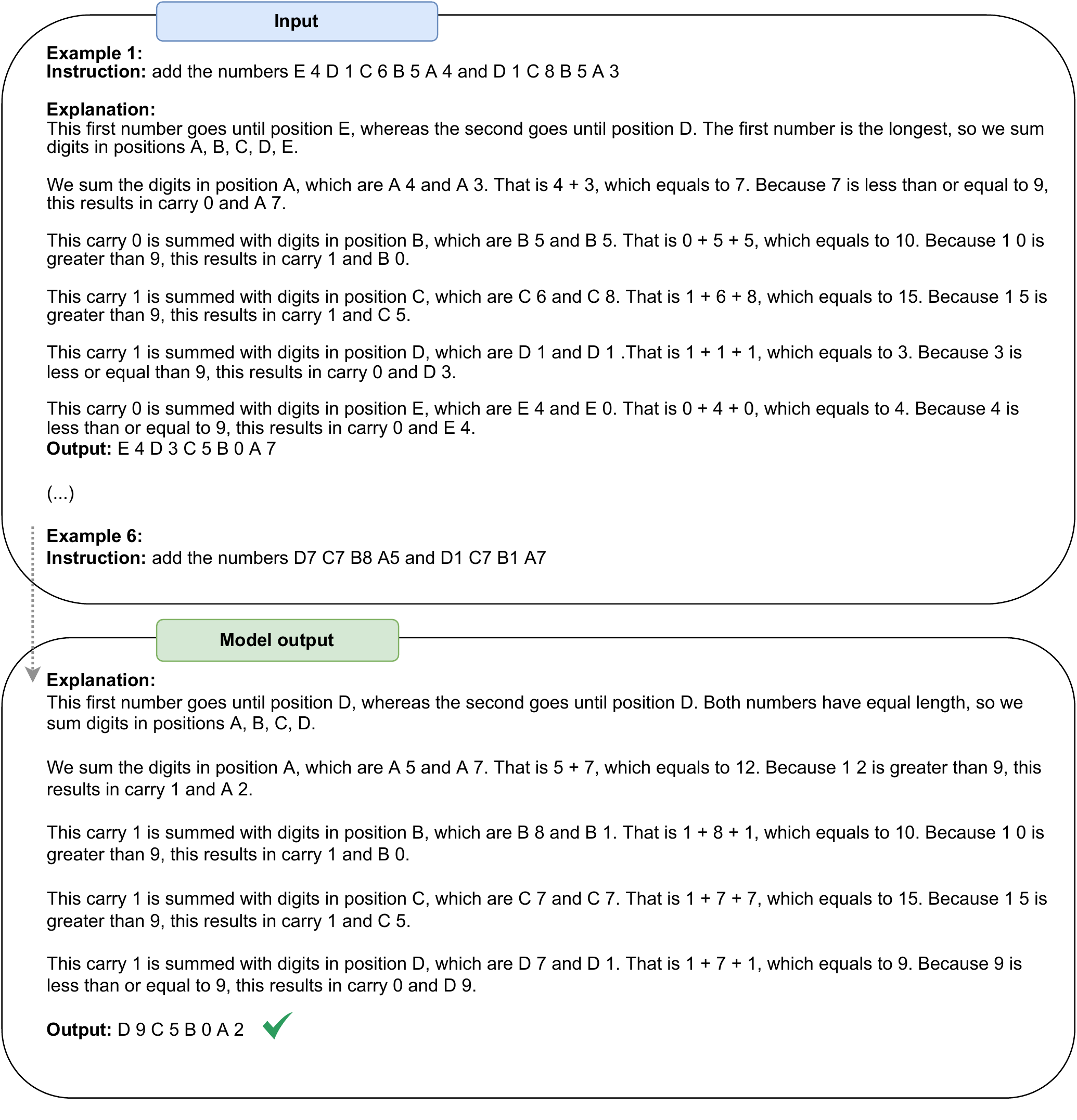}
  \caption{Example of a few-shot prompt and model completion for the addition task. First, a prompt composed of in-context ($ic$) samples are given, which are formed by $\{input, explanation, output\}$ triplets concatenated with an out-of-context ($oc$) test example that has only the "instruction" field. The model then completes the ``explanation'' and ``output'' fields from the test example as a result.}
  \label{fig:prompt-example}
\end{figure}

In this section, we describe our proposed method for inducing explanations and markup tokens using in-context learning with a few examples.
We first create a prompt $ic||oc$ that concatenates in-context training examples $ic$ with a test example $oc$. The $ic$ examples consist of $N$ triples of ``Instruction'', ``Explanation'' and ``Output'', i.e., $ic=\{(i_1^*,e_1^*,o_1^*), ..., (i_N^*,e_N^*,o_N^*)\}$. 
The test example $oc$ is made of only the ``Instruction'' field. When we feed $ic||oc$ to a language model, it should generate the remaining ``Explanation'' and ``Output'' fields for $oc$. Figure~\ref{fig:prompt-example} illustrates the input prompt given to the model and the (correct) output given by the model.

We also interleave the tokens $ic$ and $oc$ with markup tokens that help the model to precisely identify the tokens in the input and output sequences (see Figure~\ref{fig:overview}-(b) for an example). These tokens support the model in three ways: 1) They act as a form of working memory to indicate progress being made. 2) They act as sub-prompt anchors to inform the start of a known pattern. 3) They implicitly model a stopping condition should a certain amount of progress be reached.
We programmatically include these markups in each test input and remove them from the output answers before comparing them with ground-truth ones.

Due to its few-shot nature, our method can be adjusted for different tasks. Likewise, our approach does not require any additional modifications to the language model such as pretraining or changes to the loss function.

\section{Experimental Setup}
We evaluate our method in two tasks that require extrapolation: 1) the length split of the SCAN data~\cite{lake2018generalization}  and 2) the addition of two numbers. In all experiments, we used the \texttt{text-davinci-002} model, available via a paid API provided by OpenAI. 
We report the accuracy of the test set.

\subsection{SCAN}
The SCAN synthetic dataset translates simple navigation commands into a sequence of actions (e.g., the input \texttt{jump thrice} results in the output \texttt{JUMP JUMP JUMP}). These commands are generated from the composition of a specific grammar, combining ``primitive'' commands such as \texttt{jump, walk, look, run and turn}; ``modifiers'' (\texttt{left},  \texttt{right}, \texttt{around}, \texttt{opposite}); repetition symbols like \texttt{twice}/\texttt{thrice}; ``combiners'' (\texttt{and}/\texttt{after}) that group two action sequences.

To construct the prompt, we generated nine in-context training examples, each made of three parts: an instruction, an explanation, and the desired output. The ``Instruction'' is a sequence of commands while the ``Explanation'' is a description, in natural language detailing the steps to generate the output. The ``Output'' corresponds to the expected answer to the instruction. In addition, in the output field, we inject markup tokens to delimit the end of a repeating sequence or sub-instruction. Therefore, to indicate each repetition of a given action, we use positive integers and at the end of a sequence of actions, we use the separator \texttt{||}. For example, for the input: \texttt{jump twice and walk twice}, we generate the output \texttt{JUMP 1 JUMP 2 || WALK 1 WALK 2}.

The target outputs of training examples have up to 22 actions.
The test examples were drawn from the ``Length'' split provided by the authors.\footnote{https://github.com/brendenlake/SCAN}
This set has 3,920 examples whose target output varies between 24 and 48 actions. The instruction (input) of each test example is appended to the in-context training examples and the model is prompted to generate the ``Explanation'' and ``Output'' fields.
Thus, since training examples are shorter than test ones, we are able to assess the compositional generalization of the model while extrapolating to larger unseen sequences.
Due to the cost of using the GPT-3 API (approximately 0.10 USD per example), we evaluated the model on 400 randomly sampled examples from the test set.





\subsection{Addition Task}
Extrapolation abilities can also be tested with arithmetic tasks. For this, we built a prompt for the addition operation, where we present five in-context training examples with two numbers up to 5 digits and ask the model to generate the explanation and answer for a test set example made of numbers with 4 to 14 digits. We evaluate the model on 400 test samples automatically generated by the ``balanced sampling'' method from Nogueira et al.~\cite{nogueira2021investigating}, which ensures that the set will have a roughly equal proportion of answers with $d$-digit numbers, with $d \in [4, 14]$.

We use a template similar to SCAN's to feed the in-context examples to the model. We manually generate the explanations for the training examples and inject markup tokens in the instructions and the target output. In the expected output, these tokens are used during the explanation steps. 
We illustrate in Figure~\ref{fig:prompt-example} an example of a prompt followed by a completion of the model.

\section{Results}

\begin{table}

\centering

\begin{tabular}{lr}
 \toprule
 \textbf{Method} & \textbf{Acc.} \\
 \midrule
 \textit{Specialized Architectures}\\
 Syntactic Attn.~\cite{russin2019compositional} & 15.2 \\
 CGPS~\cite{li-etal-2019-compositional} & 20.3 \\
 T5-base DUEL~\cite{zhu2021learning} & 45.0 \\
 LANE~\cite{liu2020compositional} & 100.0 \\
 NSSM~\cite{chen2020compositional} & 100.0 \\
 SBSP~\cite{herzig-berant-2021-span} & 100.0 \\
 NQG~\cite{shaw2021compositional} & 100.0 \\
 Synth~\cite{nye2020learning} & 100.0 \\
\midrule
 \textit{General-purpose Architectures}\\
 T5-base~\cite{furrer2020compositional} & 14.4\\
 T5-Large~\cite{furrer2020compositional} & 5.2 \\
 T5-3B~\cite{furrer2020compositional} & 3.3\\
 T5-11B~\cite{furrer2020compositional} & 2.0 \\
 GPT-3 Ada - fine-tuned & 13.9 \\
 GPT-3 Curie - fine-tuned & 6.4 \\
 GPT-3 Davinci - fine-tuned & 8.2 \\
 Least-to-Most~\cite{zhou2022least} & 99.7 \\
 ---\\
 Ours (rationales only) & 2.5\\
 Ours (markups only) & 22.5\\
 Ours (rationales + markupts - inverted prompt) & 30.0 \\
 Ours (rationales + markups) & 95.2 \\
 \bottomrule
\end{tabular}
\caption{Results on the ``length'' split of the SCAN dataset. }
\label{tab:results_scan}
\end{table}

\begin{figure}
\centering
\begin{tikzpicture}[scale = 0.9]
\begin{axis}[
    legend cell align=left,
    legend style={font=\small},
    legend style={at={(0.97,0.43)}},
    xlabel=Digits in the ground-truth answer,
    ylabel=Test Acurracy (\%),
    xmin=4, xmax=14,
    ymin=0, ymax=100,
    xtick={4,5,6,7,8,9,10,11,12,13,14},
    xticklabels={4,5,6,7,8,9,10,11,12,13,14}, 
    ytick={0,20,40,60,80,100},
    xtick style={draw=none},
    ytick style={draw=none},
    every axis plot/.append style={thick}
    ]

\addplot[mark=square*,solid,orange] plot coordinates {
    (4,100.0)
    (5,100.0)
    (6,29.7)
    (7,0.0)
    (8,0.0)
    (9,0.0)
    (10,0.0)
    (11,0.0)
    (12,0.0)
    (13,0.0)
    (14,0.0)
};
\addlegendentry{T5 + 100k examples}

\addplot[mark=*,blue] plot coordinates {
    (4,89.47)
    (5,90.38)
    (6,82.75)
    (7,72.97)
    (8,69.04)
    (9,81.08)
    (10,84.61)
    (11,69.23)
    (12,48.64)
    (13,54.83)
    (14,60.52)
};
\addlegendentry{Ours, rationale+markup}
\addplot[mark=*,red] plot coordinates {
    (4,14.28)
    (5,12.5)
    (6,5.88)
    (7,4.34)
    (8,5.55)
    (9,0.0)
    (10,0.0)
    (11,0.0)
    (12,0.0)
    (13,0.0)
    (14,0.0)
};
\addlegendentry{Ours, markup-only}
\addplot[mark=*,black] plot coordinates {
    (4,0.0)
    (5,0.0)
    (6,0.0)
    (7,0.0)
    (8,0.0)
    (9,0.0)
    (10,0.0)
    (11,0.0)
    (12,0.0)
    (13,0.0)
    (14,0.0)
};
\addlegendentry{Ours, rationale-only}

\draw[draw = gray!40!white] (4,0) -- (4,100);  
\draw[draw = gray!40!white] (6,0) -- (6,100);  
\draw[draw = gray!40!white] (8,0) -- (8,100);  
\draw[draw = gray!40!white] (10,0) -- (10,100);  
\draw[draw = gray!40!white] (12,0) -- (12,100);  
\draw[draw = gray!40!white] (14,0) -- (14,100);  

\draw[draw = gray!40!white] (0,80) -- (14,80);  
\draw[draw = gray!40!white] (0,60) -- (14,60);  
\draw[draw = gray!40!white] (0,40) -- (14,40);  
\draw[draw = gray!40!white] (0,20) -- (14,20);  
\end{axis}
\end{tikzpicture}
\caption{Test set accuracy in the addition task vs number of digits in the ground-truth answer.} \label{fig:arthimetc-test-set}
\end{figure}

In Table~\ref{tab:results_scan} we show the results for the length split of the SCAN dataset. We see that specialized models like LANE, NSSM, and SBSP solve the compositional generalization proposed by SCAN, whereas generic architectures such as T5~\cite{raffel2020exploring} or GPT-3 \cite{brown2020language} fine-tuned on the task have poor performance.

We also show results for GPT-3's Ada (300M parameters), Curie (6B parameters) and Davinci (175B parameters) models fine-tuned on all 16,990 training examples of the SCAN dataset for 3 epochs. In these cases, we do not use in-context examples, explanations, or markup tokens.
Our methodology of providing prompts with detailed explanations was shown to be more effective than finetuning on thousands of examples.

The same behavior is also observed in the addition task, as seen in Figure~\ref{fig:arthimetc-test-set}. Our approach with explanations and markup tokens (rationale + markup) shows that even with as few as 5 examples, the model can perform the task of adding numbers with more than 5 digits, reaching a performance of around 60\% in numbers with up to 14 digits and an average accuracy of 73\% considering all 400 examples in the test set.

We also investigated the performance of fine-tuning a general-purpose model on this task. We trained a T5-base with 100K samples on numbers with 2 to 5 digits per 10 epochs without adding explanations. We observe that the model reaches 100\% accuracy with numbers of up to 5 digits, but fails to add numbers with more than 6 digits.

\subsection{Ablation: Rationale-only vs. markup-only}

We also investigate the impact of using explanation and markup tokens in isolation. We compare two scenarios: prompts without explanation (markups-only) and without markup tokens (rationales-only). 

In Table~\ref{tab:results_scan}, we see that the rationale-only and markup-only approaches have significantly lower test accuracy, demonstrating that it is not enough to explain how to solve the task, but it is also important to inject markup tokens. We believe that these tokens help the model generate repeated sequences of tokens. 

\begin{figure}
  \includegraphics[width=0.49\textwidth]{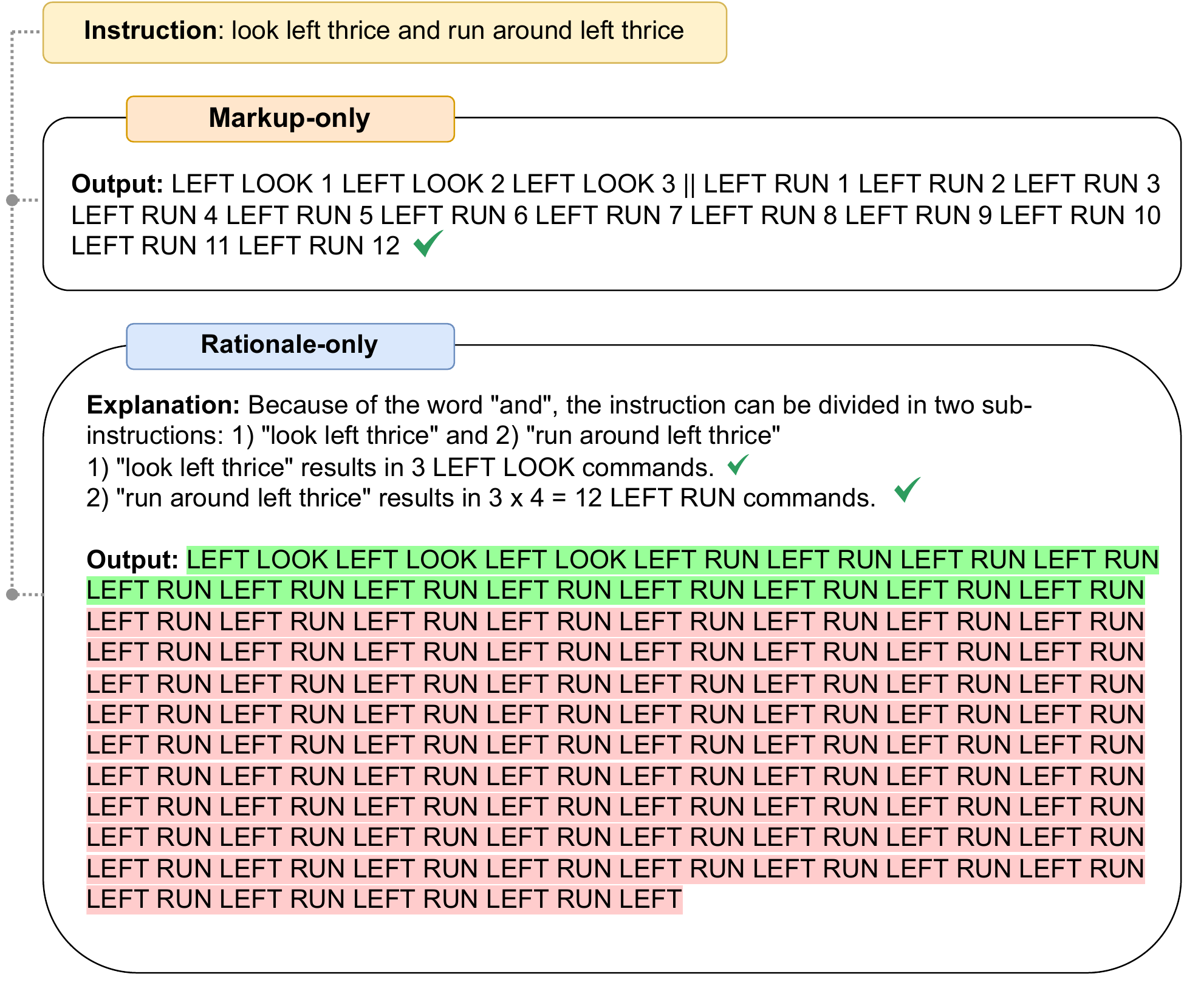}
  \caption{Model output differences between the markup-only and rationale-only approaches.}
  \label{fig:markups-vs-rationales-only}
\end{figure}

In Figure~\ref{fig:markups-vs-rationales-only},  we provide qualitative evidence of this hypothesis: Without markup tokens, the model correctly generates the explanation but fails to finish the action sequence, therefore entering a loop.

\subsection{Ablation: Inverted prompt}
We also experimented with reversing the order in which the "explanations" and "outputs" fields are presented to the model. Therefore we provide the expected output first and then the explanation. The idea of this experiment was to verify if the order explanation followed by the output has an impact on the generation of the answer. In Table~\ref{tab:results_scan} we see that the performance drops from 95.2 to 30\% (rationales + markups - inverted prompt).
This empirical result agrees with the literature in terms that the model possibly processes the explanation before determining the final output.

\section{Conclusion}
In this work, we show how step-by-step rationales and positional markup tokens enable general-purpose architectures to extrapolate to sequences that are significantly longer than those provided as training examples. Rationales before the answer break down the problem into small executable chunks and markup tokens track the working progress as the output is generated. Importantly, we show how these methods are complementary and, when used together, enable remarkable extrapolation results on two synthetic tasks.

\textit{However, we note the use of markup tokens as a limitation of current models and subword tokenizers. Future models should be able to count tokens and keep track of individual tokens in long sequences without resorting to additional supporting tokens.} 
As our qualitative analysis shows, most failure cases are due to one or two tokens generated incorrectly. We see the ability to automatically verify these errors, as proposed by Cobbe et al.~\cite{cobbe2021training}, as a promising direction to improve the extrapolation capabilities of current models.

\section*{Acknowledgments}
This research was partially funded by grants 2020/09753-5 and 2022/01640-2 from Fundação de Amparo à Pesquisa do Estado de São Paulo (FAPESP).
We also would like to thank Google Cloud for credits to support this work.
R Lotufo is partially supported by CNPq (The Brazilian National Council for Scientific and Technological Development) under grant 310828/2018-0.

\bibliography{main}
\bibliographystyle{aaai}

\appendix
\clearpage
\newpage

\section{Qualitative Analysis}
\label{sec:qualitative}
In this section, we show some correct outputs generated by our model as well as failure cases.

In Tables~\ref{tab:scan-samples-correct} and \ref{tab:scan-samples-incorrect}, we present two model completions (in light yellow background) for the test set examples from the SCAN dataset. In Table~\ref{tab:scan-samples-correct}, the model generates a correct step-by-step explanation and also correctly generates the output. In Table~\ref{tab:scan-samples-incorrect}, the model fails to construct the explanation of the second sub-instruction 'run right thrice' because it did not insert the 'RIGHT' modifier along with the primitive command 'RUN'. This mistake results in a output sequence of actions that is wrong.

Tables~\ref{tab:arthimetic-sample-correct}, \ref{tab:arthimetic-incorrect-carry}, and \ref{tab:arthimetic-incorrect-position-token} show qualitative results for the addition task. In Table~\ref{tab:arthimetic-sample-correct}, the model correctly generates the explanation and the answer. However, in Table~\ref{tab:arthimetic-incorrect-carry}, the model erroneously adds one more step, ``\textit{This remaining carry results in F 1}''. In the example of Table~\ref{tab:arthimetic-incorrect-position-token}, the model fails to obtain the digits of the position token 'E', leading to wrong outputs in the following steps. In both cases, we observe that explanations also make it easy to identify the logical flaws committed by the model.

\begin{table*}
\centering\centering\resizebox{1.0\textwidth}{!}{
\begin{tabular}{lr}
 \hline
 \begin{tabular}{lcc}
 \\
 \textcolor{blue}{(In-context examples)} \\
 \\
\begin{tabularx}{\linewidth}{X}
(...) \\
\\
Example 9: \\
Instruction: turn around right twice \\
Explanation: \\
1) "turn around right twice". The word "around" results in 4 repetitions, therefore, "turn around right" corresponds to 4 RIGHT commands. \\
Because of the word "twice", "turn around left twice" results in 2 x 4 = 8 RIGHT commands.\\
Output: RIGHT 1 RIGHT 2 RIGHT 3 RIGHT 4 RIGHT 5 RIGHT 6 RIGHT 7 RIGHT 8
\\
\\
Example 10: \\
Instruction: run around right and look around right twice \\
Explanation: \\
\colorbox{yellow!15!white}{Because of the word "and", the instruction can be divided in two sub-instructions:} \\
\colorbox{yellow!15!white}{1) "run around right" and 2) "look around right twice"} \\
\colorbox{yellow!15!white}{1) "run around right". The word "around" results in 4 repetitions, therefore,}
\colorbox{yellow!15!white}{"run around right" results in 4 RIGHT RUN commands.}\\
\colorbox{yellow!15!white}{2) "look around right twice". The word "around" results in 4 repetitions, therefore, } \\
\colorbox{yellow!15!white}{"look around right" corresponds to 4 RIGHT LOOK commands.}\\ 
\colorbox{yellow!15!white}{Because of the word "twice", "look around right twice"} \\
\colorbox{yellow!15!white}{results in 2 x 4 = 8 RIGHT LOOK commands.} \\
\colorbox{yellow!15!white}{\parbox{1\textwidth} {Output: \colorbox{green!60!white}{RIGHT RUN 1} \colorbox{green!60!white}{RIGHT RUN 2} \colorbox{green!60!white}{RIGHT RUN 3} \colorbox{green!60!white}{RIGHT RUN 4} \colorbox{green!60!white}{ \verb@||@} \colorbox{green!60!white}{RIGHT LOOK 1} \colorbox{green!60!white}{RIGHT LOOK 2} \colorbox{green!60!white}{RIGHT LOOK 3} \colorbox{green!60!white}{RIGHT LOOK 4} \colorbox{green!60!white}{RIGHT LOOK 5} \colorbox{green!60!white}{RIGHT LOOK 6}
\colorbox{green!60!white}{RIGHT LOOK 7}  \colorbox{green!60!white}{RIGHT LOOK 8}}}
\\
\end{tabularx}\\
\end{tabular}\\
\end{tabular}
}
\vspace{0.1cm}
\caption{A correct completion produced by our model in the SCAN task.}\label{tab:scan-samples-correct}
\end{table*}


\begin{table*}
\centering\centering\resizebox{1.0\textwidth}{!}{
\begin{tabular}{lr}
 \hline
 \begin{tabular}{lcc}
 \\
 \textcolor{blue}{(In-context examples)} \\
 \\
\begin{tabularx}{\linewidth}{X}
(...) \\
\\
Example 9: \\
Instruction: turn around right twice \\
Explanation: \\
1) "turn around right twice". The word "around" results in 4 repetitions, therefore, "turn around right" corresponds to 4 RIGHT commands. \\
Because of the word "twice", "turn around left twice" results in 2 x 4 = 8 RIGHT commands.\\
Output: RIGHT 1 RIGHT 2 RIGHT 3 RIGHT 4 RIGHT 5 RIGHT 6 RIGHT 7 RIGHT 8
\\
\\
Example 10: \\
Instruction: run right thrice after look around left thrice \\
Explanation: \\
\colorbox{yellow!15!white}{The word "after" results in the inversion of the two sub-instructions:} \\
\colorbox{yellow!15!white}{2) "run right thrice" and 1) "look around left thrice". } \\
\colorbox{yellow!15!white}{1) "look around left thrice". The word "around" results in 4 repetitions, therefore, }
\colorbox{yellow!15!white}{"look around left" results in 4 LEFT LOOK commands.}\\
\colorbox{yellow!15!white}{Because of the word "thrice", "look around left thrice"}
\colorbox{yellow!15!white}{results in 3 x 4 = 12 LEFT LOOK commands.}\\
\colorbox{yellow!15!white}{2) "run right thrice". The word "thrice" results in 3 repetitions, therefore,}
\colorbox{yellow!15!white}{"run right thrice" results in 3  \colorbox{red!50!white}{ RUN} commands}.\\ 
\colorbox{yellow!15!white}{\parbox{1\textwidth} {Output: \colorbox{green!60!white}{LEFT LOOK 1} \colorbox{green!60!white}{LEFT LOOK 2} \colorbox{green!60!white}{LEFT LOOK 3} \colorbox{green!60!white}{LEFT LOOK 4}
\colorbox{green!60!white}{LEFT LOOK 5}
\colorbox{green!60!white}{LEFT LOOK 6} \colorbox{green!60!white}{LEFT LOOK 7} \colorbox{green!60!white}{LEFT LOOK 8}
\colorbox{green!60!white}{LEFT LOOK 9}
\colorbox{green!60!white}{LEFT LOOK 10}
\colorbox{green!60!white}{LEFT LOOK 11}
\colorbox{green!60!white}{LEFT LOOK 12}
\colorbox{green!60!white}{ \verb@||@} \colorbox{red!50!white}{RUN 1} \colorbox{red!50!white}{RUN 2}
\colorbox{red!50!white}{RUN 3}}}
\\
\end{tabularx}\\
\end{tabular}\\
\end{tabular}
}
\vspace{0.1cm}
\caption{Completion incorrectly produced by our model in the SCAN task. The error is identified in the passage: ``\textit{"run right thrice" results in 3 RUN commands.}''. The correct is ``\textit{3 RIGHT RUN commands}''.}\label{tab:scan-samples-incorrect}
\end{table*}

\begin{table*}
\centering\centering\resizebox{1.0\textwidth}{!}{
\begin{tabular}{lr}
 \hline
 \begin{tabular}{lcc}
 \\
 \textcolor{blue}{(In-context examples)} \\
 \\
\begin{tabularx}{\linewidth}{X}
(...) \\
\\
Example 5: \\
Instruction: add the numbers C 7 B 8 A 0 and D 2 C 3 B 0 A 8 \\
Explanation: \\
This first number goes until position C, whereas the second goes until position D. \\
The second number is the longest, so we sum digits in positions A, B, C, D. \\
We sum the digits in position A, which are A 0 and A 8. That is 0 + 8, which equals to 8. Because 8 is less than or equal to 9, this results in carry 0 and A 8. \\
This carry 0 is summed with digits in position B, which are B 8 and B 0. That is 0 + 8 + 0, which equals to 8. Because 8 is less than or equal to 9, this results in carry 0 and B 8. \\
This carry 0 is summed with digits in position C, which are C 7 and C 3. That is 0 + 7 + 3, which equals to 10. Because 1 0 is greater than 9, this results in carry 1 and C 0. \\
This carry 1 is summed with digits in position D, which are D 0 and D 2. That is 1 + 0 + 2, which equals to 3. Because 3 is less than or equal to 9, this results in carry 0 and D 3. \\
Output: D 3 C 0 B 8 A 8
\\
\\
Example 6: \\
Instruction: add the numbers K 2 J 0 I 7 H 9 G 6 F 4 E 7 D 5 C 1 B 6 A 2 and K 9 J 2 I 1 H 7 G 5 F 6 E 4 D 4 C 6 B 1 A 4 \\
Explanation: \\
\colorbox{yellow!15!white}{\parbox{1\textwidth} {This first number goes until position K, whereas the second goes until position K. Both numbers have equal length, so we sum digits in positions A, B, C, D, E, F, G, H, I, J, K. \\
We sum the digits in position A, which are A 2 and A 4. That is 2 + 4, which equals to 6. Because 6 is less than or equal to 9, this results in carry 0 and A 6. \\
This carry 0 is summed with digits in position B, which are B 6 and B 1. That is 0 + 6 + 1, which equals to 7. Because 7 is less than or equal to 9, this results in carry 0 and B 7. \\
This carry 0 is summed with digits in position C, which are C 1 and C 6. That is 0 + 1 + 6, which equals to 7. Because 7 is less than or equal to 9, this results in carry 0 and C 7. \\
This carry 0 is summed with digits in position D, which are D 5 and D 4. That is 0 + 5 + 4, which equals to 9. Because 9 is less than or equal to 9, this results in carry 0 and D 9. \\
This carry 0 is summed with digits in position E, which are E 7 and E 4. That is 0 + 7 + 4, which equals to 11. Because 1 1 is greater than 9, this results in carry 1 and E 1. \\
This carry 1 is summed with digits in position F, which are F 4 and F 6. That is 1 + 4 + 6, which equals to 11. Because 1 1 is greater than 9, this results in carry 1 and F 1. \\
This carry 1 is summed with digits in position G, which are G 6 and G 5. That is 1 + 6 + 5, which equals to 12. Because 1 2 is greater than 9, this results in carry 1 and G 2. \\
This carry 1 is summed with digits in position H, which are H 9 and H 7. That is 1 + 9 + 7, which equals to 17. Because 1 7 is greater than 9, this results in carry 1 and H 7. \\
This carry 1 is summed with digits in position I, which are I 7 and I 1. That is 1 + 7 + 1, which equals to 9. Because 9 is less than or equal to 9, this results in carry 0 and I 9. \\
This carry 0 is summed with digits in position J, which are J 0 and J 2. That is 0 + 0 + 2, which equals to 2. Because 2 is less than or equal to 9, this results in carry 0 and J 2. \\
This carry 0 is summed with digits in position K, which are K 2 and K 9. That is 0 + 2 + 9, which equals to 11. Because 1 1 is greater than 9, this results in carry 1 and K 1. \\
This remaining carry results in L 1.}}

\colorbox{yellow!15!white}{\parbox{1\textwidth} {Output:\colorbox{green!60!white}{L 1 K 1 J 2 I 9 H 7 G 2 F 1 E 1 D 9 C 7 B 7 A 6}}}
\\
\end{tabularx}\\
\end{tabular}\\
\end{tabular}
}
\vspace{0.1cm}
\caption{Completion correctly produced by our model in the addition task. }\label{tab:arthimetic-sample-correct}
\end{table*}

\begin{table*}
\centering\centering\resizebox{1.0\textwidth}{!}{
\begin{tabular}{lr}
 \hline
 \begin{tabular}{lcc}
 \\
 \textcolor{blue}{(In-context examples)} \\
 \\
\begin{tabularx}{\linewidth}{X}
(...) \\
\\
Example 5: \\
Instruction: add the numbers C 7 B 8 A 0 and D 2 C 3 B 0 A 8 \\
Explanation: \\
This first number goes until position C, whereas the second goes until position D. \\
The second number is the longest, so we sum digits in positions A, B, C, D. \\
We sum the digits in position A, which are A 0 and A 8. That is 0 + 8, which equals to 8. Because 8 is less than or equal to 9, this results in carry 0 and A 8. \\
This carry 0 is summed with digits in position B, which are B 8 and B 0. That is 0 + 8 + 0, which equals to 8. Because 8 is less than or equal to 9, this results in carry 0 and B 8. \\
This carry 0 is summed with digits in position C, which are C 7 and C 3. That is 0 + 7 + 3, which equals to 10. Because 1 0 is greater than 9, this results in carry 1 and C 0. \\
This carry 1 is summed with digits in position D, which are D 0 and D 2. That is 1 + 0 + 2, which equals to 3. Because 3 is less than or equal to 9, this results in carry 0 and D 3. \\
Output: D 3 C 0 B 8 A 8
\\
\\
Example 6: \\
Instruction: add the numbers D 4 C 7 B 9 A 2 and  E 3 D 7 C 5 B 9 A 9 \\
Explanation: \\
\colorbox{yellow!15!white}{\parbox{1\textwidth} {This first number goes until position D, whereas the second goes until position E. The second number is the longest, so we sum digits in positions A, B, C, D, E. \\
We sum the digits in position A, which are A 2 and A 9. That is 2 + 9, which equals to 11. Because 1 1 is greater than 9, this results in carry 1 and A 1.\\
This carry 1 is summed with digits in position B, which are B 9 and B 9. That is 1 + 9 + 9, which equals to 19. Because 1 9 is greater than 9, this results in carry 1 and B 9. \\
This carry 1 is summed with digits in position C, which are C 7 and C 5. That is 1 + 7 + 5, which equals to 13. Because 1 3 is greater than 9, this results in carry 1 and C 3. \\
This carry 1 is summed with digits in position D, which are D 4 and D 7. That is 1 + 4 + 7, which equals to 12. Because 1 2 is greater than 9, this results in carry 1 and D 2. \\
This carry 1 is summed with digits in position E, which are E 0 and E 3. That is 1 + 0 + 3, which equals to 4. Because 4 is less than or equal to 9, this results in carry 0 and E 4. \\
\colorbox{red!50!white}{This remaining carry results in F 1.}}}

\colorbox{yellow!15!white}{\parbox{1\textwidth} {Output:\colorbox{red!50!white}{F 1} \colorbox{green!60!white}{E 4 D 2 C 3 B 9 A 1}}}
\\
\end{tabularx}\\
\end{tabular}\\
\end{tabular}
}
\vspace{0.1cm}
\caption{Completion incorrectly produced by our model in the addition task. The error occurs in the sentence ``\textit{This remaining carry results in F 1.}'', because in the previous step, the carry is equal to 0. The correct parts of the answer are colored in green, while the wrong parts are in red.}\label{tab:arthimetic-incorrect-carry}
\end{table*}


\begin{table*}
\centering\centering\resizebox{1.0\textwidth}{!}{
\begin{tabular}{lr}
 \hline
 \begin{tabular}{lcc}
 \\
 \textcolor{blue}{(In-context examples)} \\
 \\
\begin{tabularx}{\linewidth}{X}
(...) \\
\\
Example 5: \\
Instruction: add the numbers C 7 B 8 A 0 and D 2 C 3 B 0 A 8 \\
Explanation: \\
This first number goes until position C, whereas the second goes until position D. \\
The second number is the longest, so we sum digits in positions A, B, C, D. \\
We sum the digits in position A, which are A 0 and A 8. That is 0 + 8, which equals to 8. Because 8 is less than or equal to 9, this results in carry 0 and A 8. \\
This carry 0 is summed with digits in position B, which are B 8 and B 0. That is 0 + 8 + 0, which equals to 8. Because 8 is less than or equal to 9, this results in carry 0 and B 8. \\
This carry 0 is summed with digits in position C, which are C 7 and C 3. That is 0 + 7 + 3, which equals to 10. Because 1 0 is greater than 9, this results in carry 1 and C 0. \\
This carry 1 is summed with digits in position D, which are D 0 and D 2. That is 1 + 0 + 2, which equals to 3. Because 3 is less than or equal to 9, this results in carry 0 and D 3. \\
Output: D 3 C 0 B 8 A 8
\\
\\
Example 6: \\
Instruction: add the numbers H 9 G 7 F 9 E 7 D 4 C 4 B 3 A 0 and E 6 D 2 C 1 B 0 A 7 \\
Explanation: \\
\colorbox{yellow!15!white}{\parbox{1\textwidth} {This first number goes until position H, whereas the second goes until position E. The first number is the longest, so we sum digits in positions A, B, C, D, E, F, G, H. \\
We sum the digits in position A, which are A 0 and A 7. That is 0 + 7, which equals to 7. Because 7 is less than or equal to 9, this results in carry 0 and A 7. \\
This carry 0 is summed with digits in position B, which are B 3 and B 0. That is 0 + 3 + 0, which equals to 3. Because 3 is less than or equal to 9, this results in carry 0 and B 3.\\
This carry 0 is summed with digits in position C, which are C 4 and C 1. That is 0 + 4 + 1, which equals to 5. Because 5 is less than or equal to 9, this results in carry 0 and C 5.\\
This carry 0 is summed with digits in position D, which are D 4 and D 2. That is 0 + 4 + 2, which equals to 6. Because 6 is less than or equal to 9, this results in carry 0 and D 6.\\
This carry 0 is summed with digits in position E, which are \colorbox{red!50!white}{E 7 and E 0}. That is \colorbox{red!50!white}{0} + 7 + 0, which equals to 7. Because 7 is less than or equal to 9, this results in carry 0 and \colorbox{red!50!white}{E 7}.\\
This carry 0 is summed with digits in position F, which are F 9 and F 0. That is 0 + 9 + 0, which equals to 9. Because 9 is less than or equal to 9, this results in carry 0 and F 9.\\
This carry 0 is summed with digits in position G, which are G 7 and G 0. That is 0 + 7 + 0, which equals to 7. Because 7 is less than or equal to 9, this results in carry 0 and G 7. \\
This carry 0 is summed with digits in position H, which are H 9 and H 0. That is 0 + 9 + 0, which equals to 9. Because 9 is less than or equal to 9, this results in carry 0 and H 9.\\
}}

\colorbox{yellow!15!white}{\parbox{1\textwidth} {Output:\colorbox{green!60!white}{H 9} \colorbox{red!50!white}{G 7 F 9 E 7}\colorbox{green!60!white}{D 6 C 5 B 3 A 7}}}
\\
\end{tabularx}\\
\end{tabular}\\
\end{tabular}
}
\vspace{0.1cm}
\caption{Completion incorrectly produced by our model in the addition task. The error is in the sentence ``\textit{This carry 0 is summed with digits in position E, which are E 7 and E 0.}''. The correct values for position E are 7 and 6.}\label{tab:arthimetic-incorrect-position-token}
\end{table*}

\end{document}